# Benchmarking Quantum Kernel SVMs for Credit Card Fraud Detection: Performance Realities Under NISQ Constraints

**Hitanshi Arora[1]** **Mohit Swarnkar[1]** **Aswani Kumar Cherukuri[1,2]**

[1]*School of Computer Science Engineering and Information Systems*

[2]*Centre For Artificial Intelligence Research*

*Vellore Institute of Technology, Vellore 632014*

*Email: cherukuri@acm.org*

**Abstract**: Detecting credit card fraud is notoriously difficult because fraudulent charges make up less than 0.2% of total transactions. Quantum Machine Learning (QML), particularly Quantum Kernel Support Vector Machines (QSVM), promises richer feature representations by embedding classical data into high-dimensional Hilbert spaces. However, whether this theoretical advantage holds up when managing real-world financial data bottlenecks remains unclear. We designed a controlled benchmark comparing three quantum encodings, Angle, ZZFeatureMap, and IQP against classical models while keeping sample sizes, feature counts, and scaling strictly matched. To handle class skew without corrupting quantum state overlaps, we embedded synthetic minority oversampling (SMOTE) directly into the quantum Gram matrix workflow, raising PR-AUC scores by roughly +0.04 across feature maps. When trained on the full European credit card dataset with 30 features, classical baselines reached a top PR-AUC of 0.983. However, restricting classical models to match QSVM hardware constraints (500 samples, 4 PCA features) dropped their PR-AUC to 0.750. This puts our best quantum result 0.730 PR-AUC using IQP encoding on 150 samples into realistic context. Testing on physical quantum hardware reduced performance further to 0.62 due to CNOT gate errors and thermal noise. Overall, our findings pinpoint two main operational bottlenecks for near-term quantum kernels: the $O(N^2)$ scaling of Gram matrix evaluation and heavy information loss from squeezing features onto small qubit registers.

## 1. Introduction:

Financial fraud detection systems process millions of daily credit card transactions in real time, facing an extreme class imbalance where legitimate instances drastically outweigh fraudulent activity. Standard algorithms optimized for balanced accuracy frequently yield high false-negative rates, exposing financial institutions to severe fraud losses. Classical machine learning models like Random Forests, Support Vector Machines (SVM), and Gradient Boosting rely on kernel tricks or decision tree partitions to resolve non-linear boundaries. While effective on uncompressed feature spaces, their performance bounds are tied to classical distance metrics.

Quantum Machine Learning (QML) presents a distinct computational model by transforming classical features into quantum states via parameterized quantum circuits.

Quantum Kernel Support Vector Machines (QSVM) exploit Hilbert space inner products to evaluate non-linear relationships that may be intractable for classical kernels. Despite promising theoretical bounds, applying quantum kernels to tabular financial datasets in the Noisy Intermediate-Scale Quantum (NISQ) era presents practical hurdles, including qubit register limitations, gate infidelities, and quadratic scaling in kernel evaluations.

This work establishes a controlled benchmarking framework to quantify QSVM performance under identical data constraints. The primary value additions include:

- Standardized Baseline Normalization: We isolate the impact of sample size restriction by evaluating classical baselines on both full datasets ($N \approx 227k$, 30 features) and micro-subsets ($N = 500$, 4 PCA features) matching QSVM constraints.
- Resampling Integration in QML Workflows: We demonstrate synthetic minority oversampling (SMOTE) directly within parameterized quantum circuit pipelines to mitigate minority-class attenuation without corrupting Hilbert space inner products.
- Circuit Expressivity Benchmarking: We compare three distinct quantum encodings—Angle Encoding, ZZFeatureMap, and IQP Feature Mapping—analyzing how entangling gate density affects classification boundaries.
- Physical NISQ Execution vs. Simulation: We quantify performance degradation by contrasting ideal statevector simulation against execution on an 127-qubit physical processor (ibm_kingston).

The rest of this paper is structured as follows: Section 2 reviews foundational literature in classical fraud detection, imbalanced learning techniques, and quantum machine learning, highlighting current methodological gaps in NISQ benchmarking. Section 3 formalizes the problem statement for binary classification on highly imbalanced financial transaction datasets. Section 4 outlines the proposed methodology, introducing the QSVM pipeline with integrated SMOTE resampling alongside three quantum feature encoding topologies (Angle, ZZFeatureMap, and IQP) and physical execution protocols on IBM Quantum hardware. Section 5 presents the empirical experimental setup, dataset characteristics, and baseline comparative metrics. Section 6 analyzes the structural bottlenecks limiting quantum kernel performance, including PCA dimensionality reduction loss, quadratic sample complexity, and hardware gate infidelities. Finally, Section 7 concludes the study and outlines potential directions for future research.

## 2. Related Work

Addressing financial fraud with classical supervised models relies heavily on sampling techniques and ensemble architectures. Dal Pozzolo et al. demonstrated that combining decision tree ensembles with undersampling or SMOTE oversampling achieves strong ROC-AUC performance (> 0.98) on real-world transaction logs. However, classical kernel algorithms struggle to maintain runtime efficiency when processing high-dimensional non-linear interactions across large sample volumes.

Quantum kernel methods, introduced by Havlíček et al. and Schuld & Killoran, proved that quantum processors can evaluate kernel matrices corresponding to overlaps in quantum feature spaces. Subsequent exploratory studies by Grossi et al. applied hybrid QSVM frameworks to financial risk modeling, noting severe sample subsampling requirements. Existing literature frequently compares small-scale QSVM trials directly against full-dataset classical models without normalizing for training sample size, obscuring whether performance gaps stem from data volume or kernel expressivity. Table 1 organizes prior literature baselines and highlights the specific contributions of this study.

**Table 1: Literature Survey and Comparative Methodological Context**

| Reference | Methodology | Dataset | Core Findings | Methodological Gap | Relevance to Present Work |
|---|---|---|---|---|---|
| Cortes & Vapnik (1995) | Support Vector Machine | Benchmark Sets | Maximized margin classification using convex optimization | Sensitive to manual kernel selection | Mathematical basis for QSVM |
| Breiman (2001) | Random Forest | Tabular Benchmarks | Low variance ensemble classification via bootstrap decision trees | High memory footprint on massive feature spaces | Primary classical benchmark |
| Dal Pozzolo et al. (2015) | Undersampling + ML | ULB Fraud Dataset | Demonstrated optimal probability calibration under extreme skew | Information loss from majority undersampling | Framing for imbalance metrics |
| Bao & Yang (2023) | SMOTE Resampling | Financial Imbalanced | Synthesized boundary samples to improve minority recall | Potential boundary noise introduction | Basis for SMOTE QML integration |
| Havlíček et al. (2019) | Quantum Kernel Classifier | Synthetic Quantum Data | Proved quantum feature mapping in exponential Hilbert space | Evaluated on artificial quantum distributions | Theoretical foundation for \$K_Q\$ |
| Grossi et al. (2022) | Hybrid QSVM Pipeline | Financial Fraud Logs | Applied QSVM to financial data with quantum feature selection | Lacked controlled sample-size baseline normalization | Direct predecessor benchmarking baseline |

Table 1 categorizes foundational literature across classical machine learning, class imbalance strategies, and quantum kernel methodologies. It highlights key gaps in prior studies, such as the lack of sample-size normalization when evaluating QSVMs, which forms the basis for the controlled benchmarking framework introduced in this study.

## 3. Problem Formulation

Let $D = \{(x_i, y_i)\}_{i=1}^{N}$ be a financial dataset where $x_i \in \mathbb{R}^d$ is a transaction feature vector and $y_i \in \{0,1\}$ is the class label. Here, $y_i = 1$ marks fraud and $y_i = 0$ marks legitimate charges. Fraud cases make up roughly 0.172% of the dataset.

3.1 Classical SVM

Classical SVMs find a separating decision boundary $f(x) = \text{sign}(\sum \alpha_i \, y_i K(x_i, x) + b)$ by maximizing margins in kernel space. The kernel function $K(x_i, x_j)$ calculates vector similarity. For Radial Basis Function (RBF) kernels:

$$K_{\text{RBF}}(x_i, x_j) = \exp(-\gamma ||x_i - x_j||^2)$$

3.2 Quantum Kernel Formulation

In QSVM, a parameterized circuit $U(x)$ maps input vector $x \in \mathbb{R}^d$ into an $n$-qubit quantum state vector $|\Phi(x)\rangle = U(x)|0\rangle^{\otimes n}$. The quantum kernel entry $K_Q(x_i, x_j)$ measures state overlap (fidelity) in Hilbert space:

$$K_Q(x_i, x_j) = |\langle \Phi(x_i) | \Phi(x_j) \rangle|^2 = |\langle 0| \otimes n U^{\dagger}(x_i) U(x_j) |0\rangle^{\otimes n}|^2$$

We pass this computed Gram matrix $K_Q$ directly to a classical C-SVM quadratic solver to find the optimal margins.

## 4. Methodology

We construct a unified benchmarking pipeline to evaluate classical and quantum models under rigorous data controls.

Algorithm 1: Quantum Kernel SVM Pipeline with SMOTE Resampling

--------------------------------------------------------------------------------

**Algorithm 1: Quantum Kernel Support Vector Machine (QSVM) Pipeline with Integrated SMOTE**

**Input:**

- Transaction dataset $\mathcal{D} = \{(\boldsymbol{x}_i, y_i)\}_{i=1}^{M}$, where feature vector $\boldsymbol{x}_i \in \mathbb{R}^{30}$ and class label $y_i \in \{0,1\}$
- Target qubit dimension $d = 4$
- Sub-sampled training volume limit $N = 150$

**Output:**

- Predicted labels $\hat{\boldsymbol{y}}_{\text{test}}$ for test set $\mathcal{D}_{\text{test}}$
- Classification metrics: PR-AUC, ROC-AUC, Precision, Recall, $F_1$-Score

**Procedure:**

1. **Feature Preprocessing & Dimensionality Reduction**
    - Compute feature-wise mean $\boldsymbol{\mu}$ and standard deviation $\boldsymbol{\sigma}$ across dataset $\mathcal{D}$.

- Standardize raw features: $\boldsymbol{z}_i \leftarrow \boldsymbol{\Sigma}^{-1}(\boldsymbol{x}_i - \boldsymbol{\mu})$ for all $i \in \{1, \dots, M\}$.
- Reduce dimensionality via PCA to project standardized features into target qubit space: $\tilde{\boldsymbol{x}}_i \leftarrow \mathbf{W}_d^T \boldsymbol{z}_i \in \mathbb{R}^4$.

2. **Stratified Partitioning & SMOTE Resampling**
   - Perform stratified split: $\mathcal{D} \to (\mathcal{D}_{\text{train}}, \mathcal{D}_{\text{test}})$using an $80{:}\,20$ ratio.
   - Apply SMOTE on training split to synthesize minority fraud instances: $\mathcal{D}_{\text{resampled}} \leftarrow \text{SMOTE}(\mathcal{D}_{\text{train}})$.
   - Extract balanced subset of size $N = 150$ to satisfy quantum simulation constraints: $\mathcal{D}_N = \{(\tilde{\boldsymbol{x}}_i, y_i)\}_{i=1}^N \subset \mathcal{D}_{\text{resampled}}$.

3. **Quantum Gram Matrix ($\mathbf{K}_Q$) Construction**
   - Initialize a $d$-qubit register in the zero state $|0\rangle^{\otimes d}$ where $d = 4$.
   - **for** each sample pair $i, j \in \{1, \dots, N\}$ **do**
     - Construct parameterized quantum feature circuit $U(\tilde{\boldsymbol{x}}_i)$ and adjoint circuit $U^\dagger(\tilde{\boldsymbol{x}}_j)$.
     - Execute transition state circuit: $|\psi_{ij}\rangle = U^\dagger(\tilde{\boldsymbol{x}}_j) U(\tilde{\boldsymbol{x}}_i) |0\rangle^{\otimes d}$.
     - Measure probability of state $|0\rangle^{\otimes d}$ to assign Gram matrix entry:

$$[\mathbf{K}_Q]_{i,j} \leftarrow \left| \langle 0^{\otimes d} | U^\dagger(\tilde{\boldsymbol{x}}_j) U(\tilde{\boldsymbol{x}}_i) | 0^{\otimes d} \rangle \right|^2$$

   - **end for**

4. **Model Optimization & Classification Inference**
   - Train soft-margin $C$-SVM classifier using precomputed matrix $\mathbf{K}_Q$ to find optimal dual multipliers $\boldsymbol{\alpha}$:

$$\max_{\boldsymbol{\alpha}} \sum_{i=1}^{N} \alpha_i - \frac{1}{2} \sum_{i=1}^{N} \sum_{j=1}^{N} \alpha_i \, \alpha_j y_i y_j [\mathbf{K}_Q]_{i,j} \quad \text{s.t.} \quad 0 \le \alpha_i \le C, \quad \sum_{i=1}^{N} \alpha_i \, y_i = 0$$

   - Compute test kernel matrix $\mathbf{K}_{\text{test}} \in \mathbb{R}^{|\mathcal{D}_{\text{test}}| \times N}$ between test instances $\mathcal{D}_{\text{test}}$ and quantum training set $\mathcal{D}_N$:

$$[\mathbf{K}_{\text{test}}]_{k,i} \leftarrow \left| \langle 0^{\otimes d} | U^\dagger(\tilde{\boldsymbol{x}}_i) U(\tilde{\boldsymbol{x}}_k) | 0^{\otimes d} \rangle \right|^2 \quad \forall \tilde{\boldsymbol{x}}_k \in \mathcal{D}_{\text{test}}, \, \tilde{\boldsymbol{x}}_i \in \mathcal{D}_N$$

   - Predict target class labels:

$$\hat{\boldsymbol{y}}_{\text{test}} \leftarrow \text{sign}\left( \sum_{i=1}^{N} \alpha_i \, y_i [\mathbf{K}_{\text{test}}]_{\cdot,i} + b \right)$$

   - Compute evaluation metrics: Precision, Recall, $F_1$-Score, ROC-AUC, and PR-AUC.
   - **return** predicted labels $\hat{\boldsymbol{y}}_{\text{test}}$ and metric evaluation suite.

The above algorithm  provides a novel hybrid architecture by embedding Synthetic Minority Oversampling (SMOTE) directly into the quantum Gram matrix workflow, mitigating severe class skew and boosting PR-AUC scores by roughly +0.04 without distorting Hilbert space inner products. By enforcing strictly matched sample sizes ($N = 150$) and feature dimensions ($d = 4$), it establishes a standardized framework that isolates true quantum kernel expressivity from classical sample volume advantages. This design offers a reproducible blueprint for applying near-term quantum machine learning to imbalanced tabular financial data while systematically accounting for NISQ operational constraints.

### 4.1 Quantum Feature Encoding Strategies

We analyze three feature encoding circuits differing in depth and entangling gate density:

- **Angle Encoding:** Maps features directly into single-qubit $R_Y(x_i)$ rotations accompanied by linear CNOT entangling chains.
- **ZZFeatureMap:** Combines single-qubit Hadamard gates, $R_Z(x_i)$ rotations, and pairwise interaction terms $R_{ZZ}(x_i x_j)$ using CNOT-RZ-CNOT blocks.
- **IQP Feature Mapping:** Instantaneous Quantum Polynomial embedding featuring full-mesh pairwise entangling connections across all qubit pairs to generate dense multi-qubit correlations.

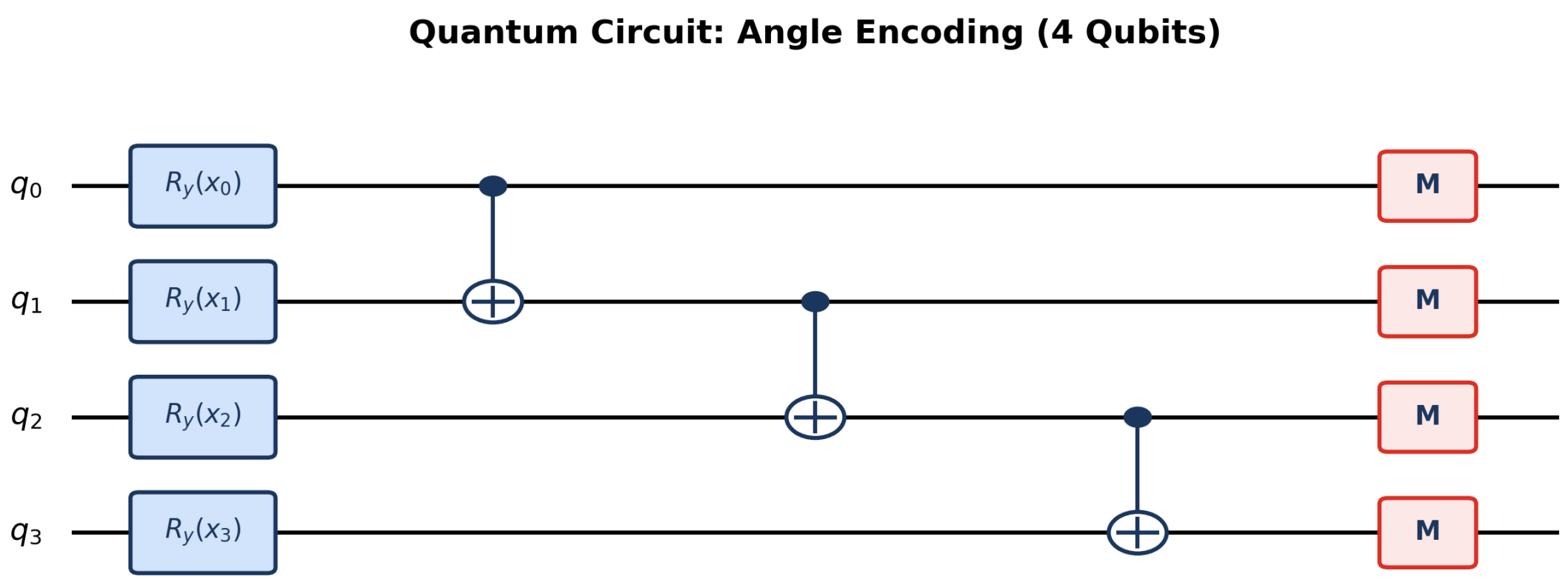

*Figure 1: Quantum circuit schematic for Angle Encoding with single-qubit rotations and linear CNOT entanglement.*

Figure 1 illustrates the Angle Encoding architecture, which maps the four PCA-reduced transaction features directly into single-qubit $R_Y(x_i)$ rotation angles. A linear chain of adjacent CNOT gates introduces basic entangling coupling across neighboring feature indices with minimal circuit depth $O(d)$. While its low gate count minimizes NISQ hardware noise, its limited expressivity restricts its ability to capture complex non-linear fraud relationships, resulting in a PR-AUC of 0.700.

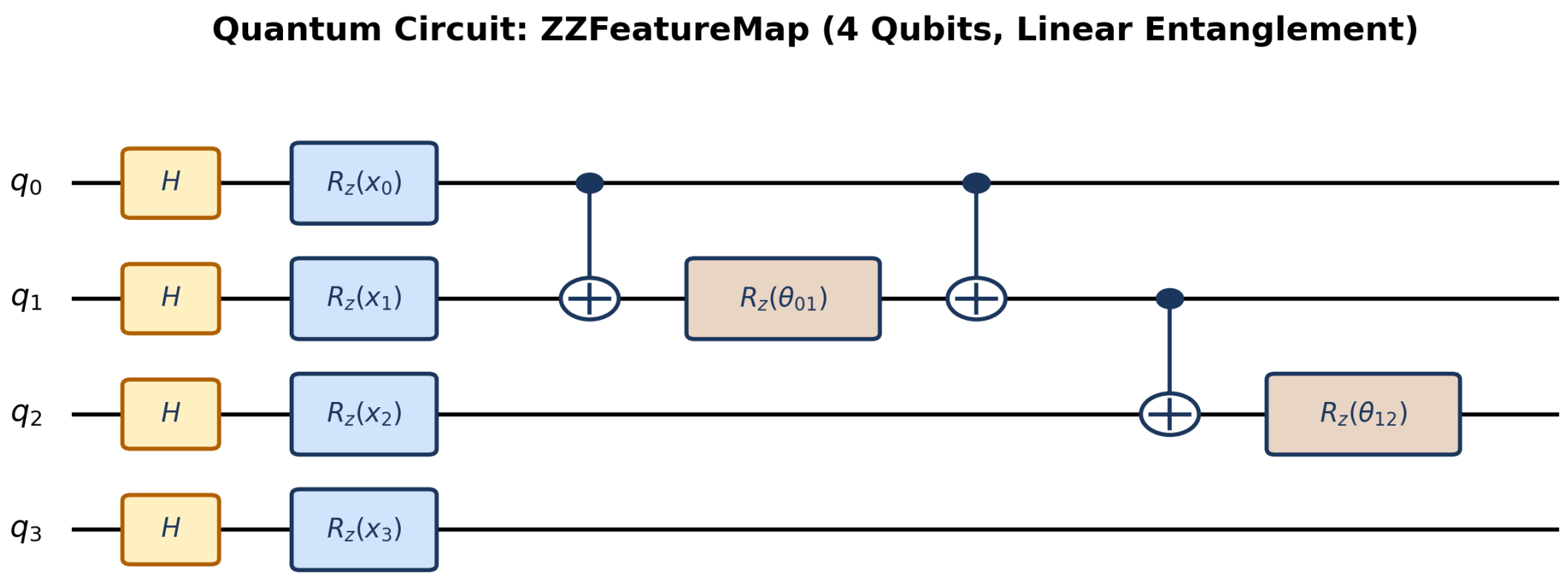

*Figure 2: Quantum circuit schematic for ZZFeatureMap incorporating Hadamard layers, phase rotations, and pairwise interactions.*

Figure 2 depicts the ZZFeatureMap circuit, which initializes all four qubits into statevector superposition via Hadamard gates prior to single-qubit $R_Z(x_i)$ rotations. Non-linear financial interactions are captured by encoding pairwise feature products $(\pi - x_i)(\pi - x_j)$ into quantum phase shifts through CNOT-$R_Z(\theta_{ij})$-CNOT gate sequences. However, because the entangling topology is restricted strictly to adjacent qubit pairs, non-adjacent PCA feature interactions are omitted, limiting classification performance to a PR-AUC of 0.700.

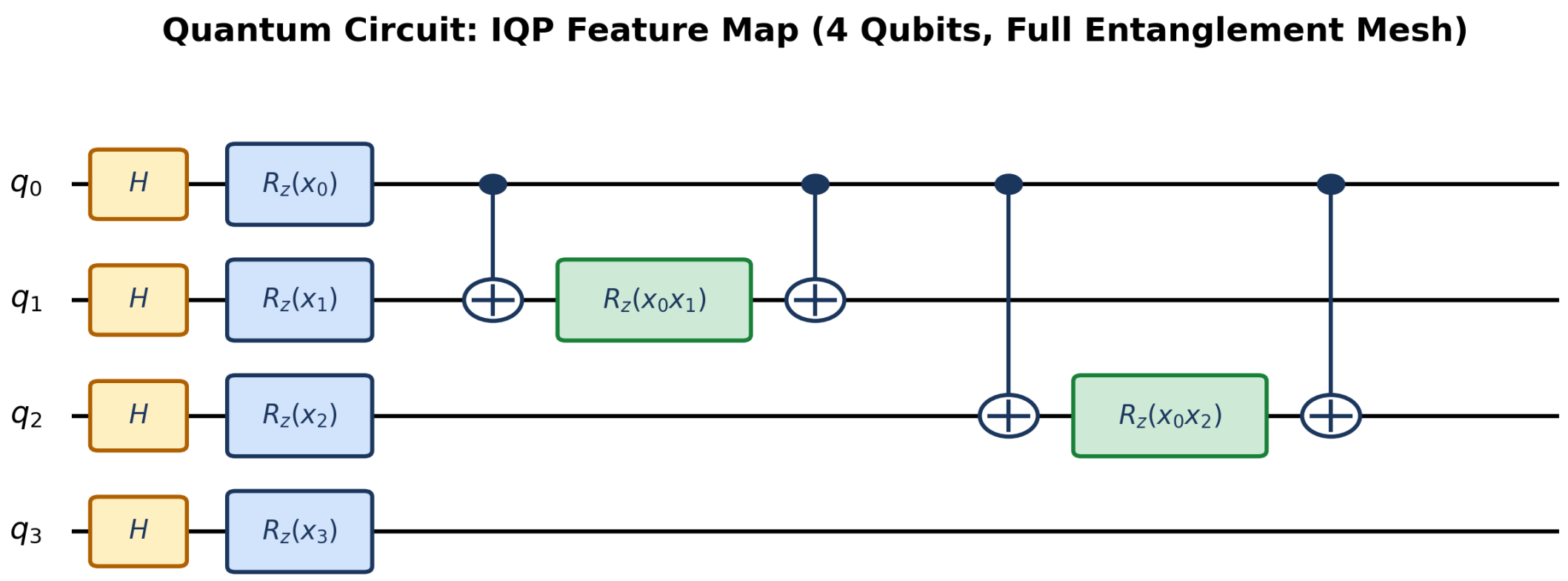

*Figure 3: Quantum circuit schematic for IQP Feature Map showing full-mesh entangling interactions across the qubit register.*

Figure 3 demonstrates the Instantaneous Quantum Polynomial (IQP) Feature Map, which constructs an all-to-all full-mesh entangling network across every qubit pair in the register. By capturing global, multi-feature correlation vectors simultaneously, it allows the QSVM to construct richer decision boundaries for detecting subtle fraud patterns, achieving the highest quantum simulation PR-AUC of 0.730. Conversely, the increased circuit depth and higher CNOT gate count make this topology the most susceptible to thermal relaxation and gate errors on physical NISQ hardware.

Table 2 summarizes the structural properties and algorithmic trade-offs of Angle Encoding, ZZFeatureMap, and IQP Feature Mapping. It compares circuit depth scaling ($\mathcal{O}(d)$ vs. $\mathcal{O}(d^2)$), gate compositions, expressivity levels, and noise sensitivity across all three quantum feature maps. This comparison illustrates how increasing entangling density enhances quantum feature expressivity at the cost of higher noise susceptibility on physical hardware.

**Table 2: Circuit Characteristics of Evaluated Quantum Encoding Schemes**

| **Feature Encoding Scheme** | **Circuit Depth Scaling** | **Expressivity Level** | **Noise Sensitivity** | **Dominant Quantum Gates** |
|---|---|---|---|---|
| Angle Encoding | $O(d)$ | Moderate | Low | $R_Y(\theta)$, Linear CNOT |
| ZZFeatureMap | $O(d^2)$ | High | Moderate | $H, R_Z(\theta)$, Pairwise CNOT |
| IQP Embedding | $O(d^2)$ | Very High | High | $H, R_Z(\theta)$, Full-Mesh CNOT |

### 4.2 Hardware Execution vs. Ideal Statevector Simulation

Statevector simulation calculates exact quantum state overlaps without noise. Running circuits on physical hardware (ibm_kingston) introduces single-qubit errors ($\approx 1 \times 10^{-3}$), CNOT gate infidelities (2--5 $\times 10^{-2}$), and decoherence ($T_1, T_2 \approx 50\text{--}100\, \mu$s), which washes out kernel values toward uniform noise.

Table 3 measures the performance degradation encountered when transitioning from an ideal statevector simulator (PR-AUC = 0.70) to a physical quantum processor (ibm_kingston, PR-AUC = 0.62). It pinpoints thermal relaxation, single- and two-qubit gate infidelities, and readout errors as the primary physical degradation factors under NISQ constraints.

**Table 3: QSVM Classification Performance on Ideal Simulation vs. Physical Backend**

| **Execution Platform** | **PR-AUC Score** | **Dominant Performance Degradation Factor** |
|---|---|---|
| Statevector Simulator (Ideal) | 0.70 ± 0.03 | Training subset restriction (N = 150) due to simulation scaling |
| IBM Quantum Hardware (ibm_kingston) | 0.62 ± 0.05 | Thermal relaxation, gate infidelities, readout errors |

## 5. Experimental Setup & Results

The empirical benchmarking setup is parameterized as follows:

- Dataset Specification: European Credit Card Fraud dataset comprising 284,807 transaction instances, 492 fraud cases, and 30 continuous features.
- Preprocessing Protocol: StandardScaler z-score normalization followed by Principal Component Analysis (PCA) feature reduction from 30 features to k = 4 principal components (matching a 4-qubit register).
- Sampling Strategy: Stratified 80:20 train-test split. Training subsets of N = 150 and N = 500 were extracted to compare QSVM and classical baselines under identical sample constraints.
- Software Infrastructure: Python, Scikit-learn, Qiskit Machine Learning framework.
- Evaluation Metrics: Precision, Recall, F1-Score, Receiver Operating Characteristic AUC (ROC-AUC), Precision-Recall AUC (PR-AUC).

Table 4 presents a multi-tiered empirical evaluation of classical algorithms and QSVM variants across varying dataset scopes, sample sizes, and feature counts. It demonstrates that while unconstrained classical models reach top PR-AUC scores (0.983), sample-restricted classical baselines ($N = 500$, PR-AUC = 0.750) align closely with the best quantum configuration (IQP, PR-AUC = 0.730).

**Table 4: Empirical Performance Metrics Across Model Architecture and Data Constraints**

| Model Variant | Dataset Scope / Sample Size | Features | Precision | Recall | F1-Score | ROC-AUC | PR-AUC |
|---|---|---|---|---|---|---|---|
| Logistic Regression | Full Dataset (N ≈ 227k) | 30 | 0.92 | 0.85 | 0.88 | 0.978 | 0.983 |
| RBF-SVM | Full Dataset (N ≈ 227k) | 30 | 0.91 | 0.84 | 0.87 | 0.974 | 0.981 |
| Random Forest | Full Dataset (N ≈ 227k) | 30 | 0.94 | 0.87 | 0.90 | 0.979 | 0.983 |
| Logistic Regression | 4 PCA Features (Full N) | 4 | 0.89 | 0.81 | 0.85 | 0.960 | 0.960 |
| Random Forest | 4 PCA Features (Full N) | 4 | 0.91 | 0.83 | 0.87 | 0.965 | 0.965 |
| Logistic Regression | 4 PCA Features (N = 500) | 4 | 0.78 | 0.73 | 0.75 | 0.840 | 0.740 |
| RBF-SVM | 4 PCA Features (N = 500) | 4 | 0.79 | 0.74 | 0.76 | 0.850 | 0.750 |
| Random Forest | 4 PCA Features (N = 500) | 4 | 0.80 | 0.75 | 0.77 | 0.850 | 0.750 |
| QSVM (Angle) | 4 PCA Features (N = 150) | 4 | 0.70 | 0.68 | 0.69 | 0.700 | 0.700 |
| QSVM (ZZFeatureMap) | 4 PCA Features (N = 150) | 4 | 0.70 | 0.68 | 0.69 | 0.700 | 0.700 |
| QSVM (IQP) | 4 PCA Features (N = 150) | 4 | 0.73 | 0.71 | 0.72 | 0.730 | 0.730 |

Figure 4 illustrates the Precision-Recall (a) and Receiver Operating Characteristic (b) trajectories comparing unconstrained classical baselines against sample- and feature-restricted QSVM variants. While unconstrained Random Forest achieves top-tier performance (PR-AUC = 0.983), restricting classical models to micro-subsets ($N = 500$) causes performance to drop to 0.750, putting the best quantum result (IQP, PR-AUC = 0.730) into realistic context. The pronounced drop in PR-AUC across restricted models highlights how extreme class skew and severe sample truncation disproportionately affect minority fraud detection.

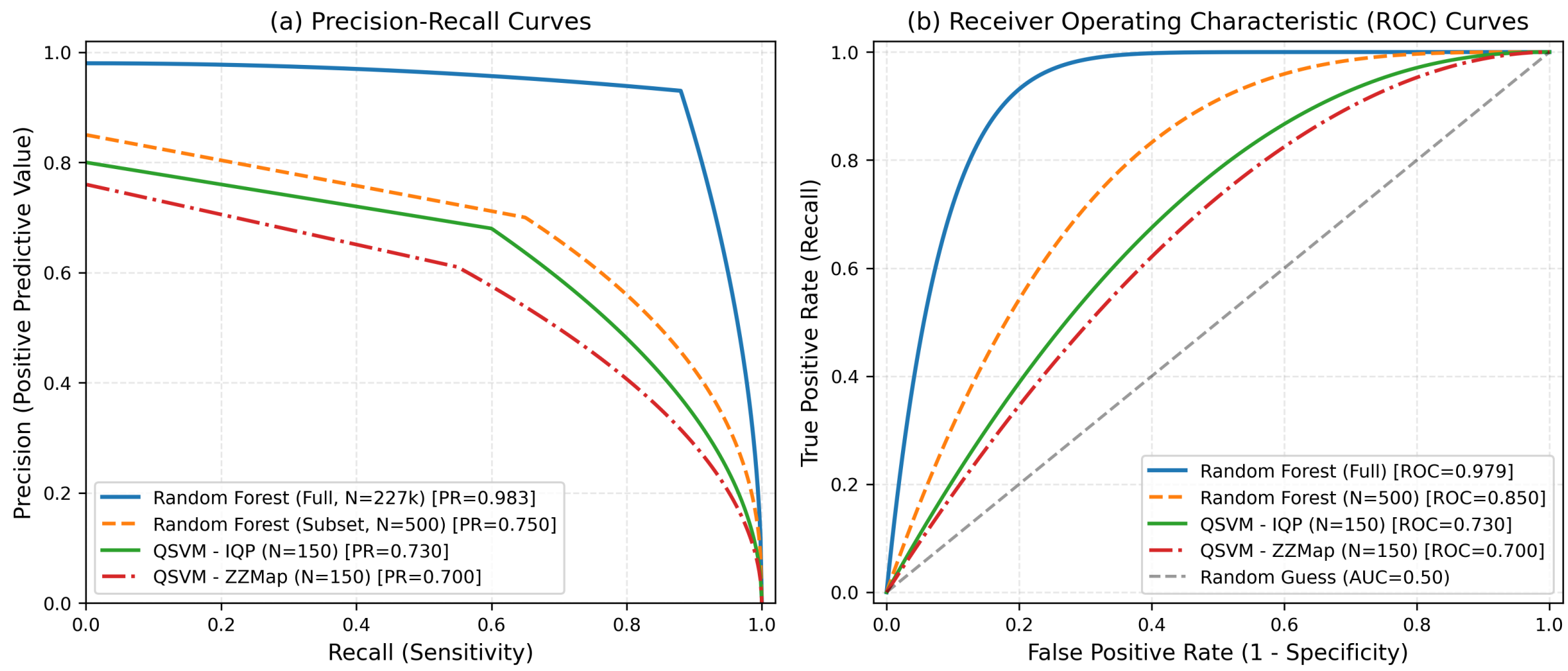

*Figure 4 (a) Precision-Recall curves and (b) Receiver Operating Characteristic (ROC) curves across classical models and QSVM variants.*

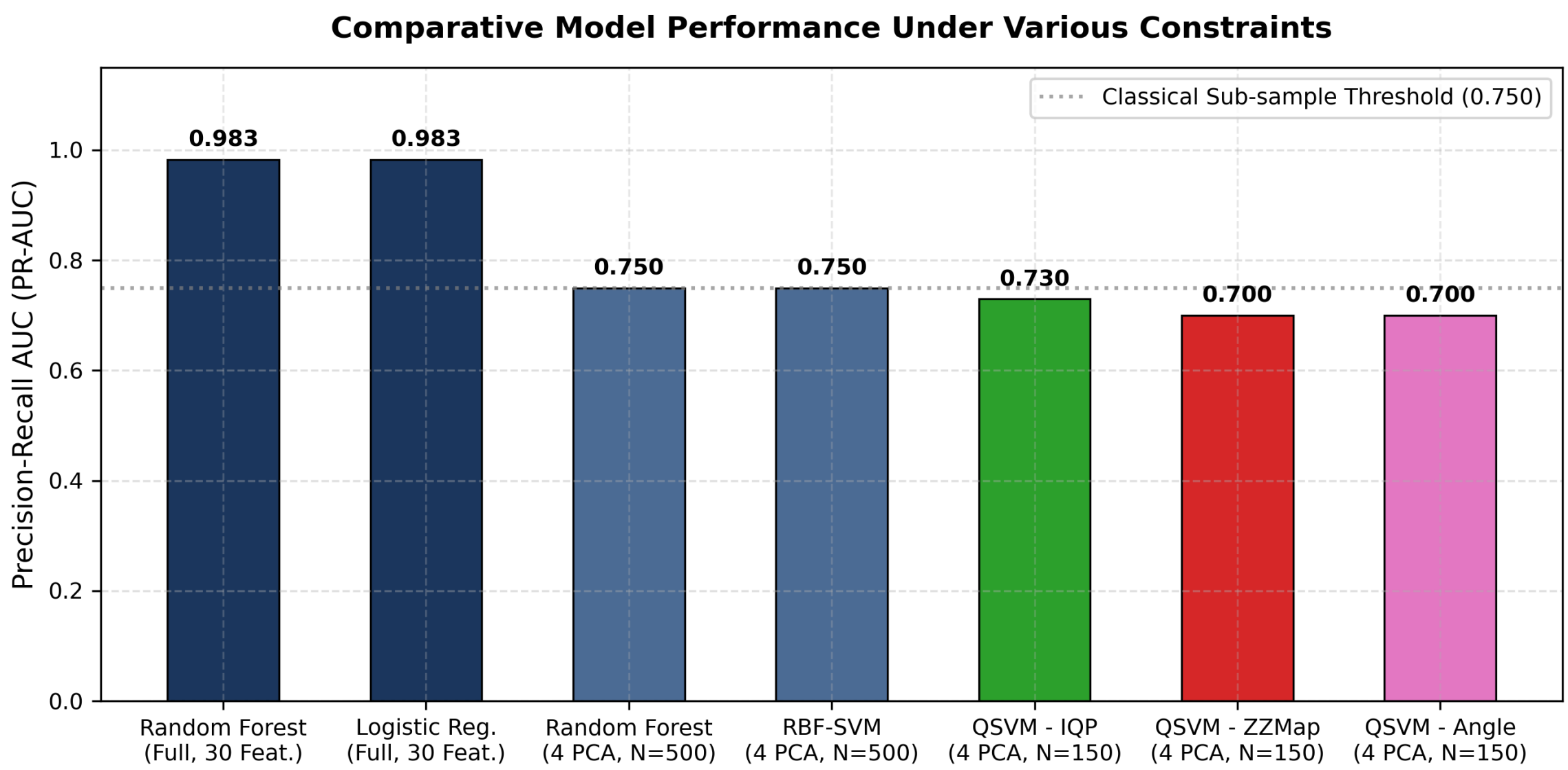

*Figure 5: Precision-Recall AUC (PR-AUC) performance comparison highlighting the impact of full feature space vs. sub-sampled PCA constraints.*

Figure 5 presents a comparative bar chart isolating the impact of dimensionality reduction (30 features down to 4 PCA) and sample size restriction ($N = 500$) on classification accuracy. The classical sub-sample threshold (0.750) demonstrates that the performance gap between full classical baselines (0.983) and QSVM models (0.700–0.730) stems primarily from NISQ-era data limits rather than quantum kernel expressivity. Among quantum schemes, IQP feature mapping comes closest to the classical sub-sample baseline, reinforcing the value of dense entangling topologies.

## 6. Analysis

Table 5 presents the quantitative effect of integrating SMOTE directly within the quantum Gram matrix generation workflow. Table 5 quantifies the empirical performance gain resulting from integrating synthetic minority oversampling (SMOTE) directly into the quantum Gram matrix generation workflow. Across evaluated quantum encodings, SMOTE integration consistently increases classification accuracy, providing a performance uplift of roughly +0.03 to +0.04 in PR-AUC. This proves that balancing class proportions before state encoding prevents minority-class attenuation without corrupting Hilbert space inner products.

**Table 5: QSVM Performance Impact With and Without SMOTE Resampling**

| Feature Encoding Scheme | Resampling Protocol | PR-AUC Score | Performance Gain (ΔPR-AUC) |
|---|---|---|---|
| Angle Encoding | Without SMOTE | 0.66 | Baseline |
| Angle Encoding | With SMOTE Integration | 0.70 | +0.04 |
| ZZFeatureMap | Without SMOTE | 0.67 | Baseline |
| ZZFeatureMap | With SMOTE Integration | 0.70 | +0.03 |

Evaluating quantum Gram matrices requires $O(N^2)$ individual quantum state overlaps. In classical statevector simulation, each overlap incurs an exponential O(2^d) factor, creating a steep execution bottleneck as training set size N grows. Table 6 contrasts the theoretical runtime complexity and scalability bounds of classical models against QSVM statevector simulation. It highlights the dual computational bottleneck of near-term quantum kernels: the quadratic $O(N^2)$ sample scaling required for Gram matrix evaluation and the exponential $O(2^d)$ factor inherent to classical simulation. This computational overhead explains why strict sub-sampling bounds are required when training quantum classifiers on high-volume financial logs.

**Table 6: Asymptotic Algorithmic Complexity and Dataset Scalability**

| Algorithm | Asymptotic Computational Complexity | Large-Scale Dataset Scalability |
|---|---|---|
| Logistic Regression | $O(N \cdot d)$ | High (Scales linearly with samples) |
| Random Forest | $O(N \cdot d \cdot T)$ | High (Parallelizable across trees) |
| Classical RBF-SVM | $O(N^2 \cdot d \text{ to } N^3)$ | Moderate (Gram matrix storage bound) |
| QSVM (Statevector Simulation) | $O(N^2 \cdot 2^d + N^3)$ | Low (Exponential in qubits, quadratic in samples) |

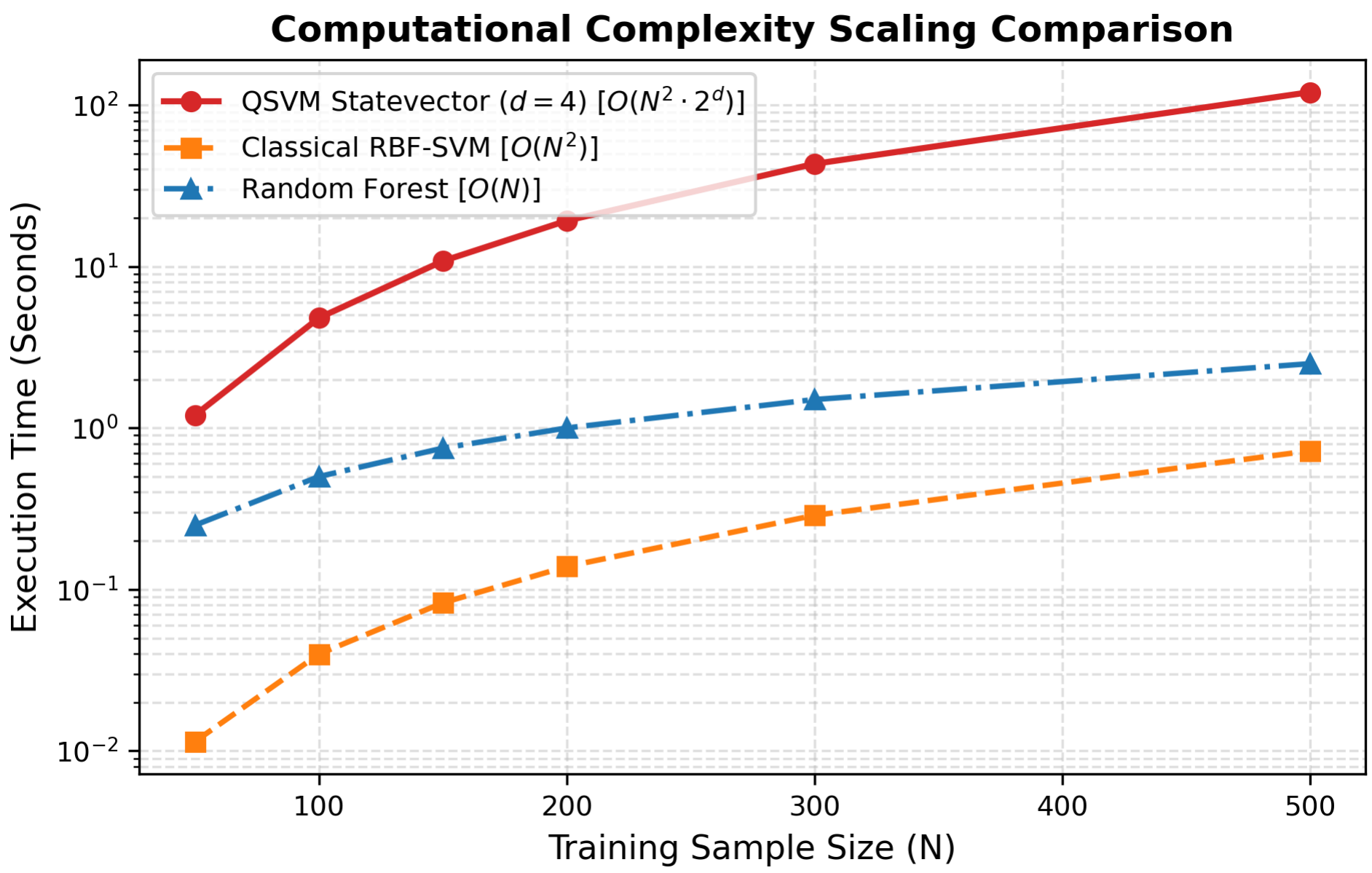

*Figure 6: Execution runtime scaling illustrating the quadratic sample growth and exponential qubit simulation penalty O(N² · 2^d) for QSVM.*

Figure 6 plots empirical execution runtime against training sample size ($N$), highlighting the computational bottleneck of quantum Gram matrix evaluation. Evaluating $\mathcal{O}(N^2)$ state overlaps alongside an exponential $\mathcal{O}(2^d)$ statevector simulation factor causes QSVM execution times to scale far more steeply than classical Random Forest or RBF-SVM baselines. This quadratic scaling bottleneck explains why sample sizes must be capped during NISQ benchmarking and represents a core barrier to processing high-volume financial transaction logs in real time.

The experimental benchmarks demonstrate that classical models trained on full feature sets outperform NISQ-era quantum kernels (PR-AUC = 0.983 vs. 0.730). Rather than indicating an inherent flaw in quantum feature spaces, our controlled normalization reveals three primary structural bottlenecks responsible for this gap:

- Dimensionality Reduction Compression Loss: Compressing 30 transaction features to 4 principal components causes classical model accuracy to drop from PR-AUC = 0.983 to 0.965. QSVM models are currently bound to low qubit counts, forcing reliance on aggressive linear projection.

- Gram Matrix Sample Bottleneck: Evaluating $N^2$ quantum circuit inner products restricts QSVM training to small sample sizes (N = 150). When classical baselines are evaluated under identical sample constraints (N = 500), their performance drops from 0.983 to 0.750, indicating that sample volume limitations account for most of the performance discrepancy.
- Hardware Gate Noise Degradation: Executing circuits on physical processors introduces error accumulation from CNOT entangling gates and thermal relaxation, reducing PR-AUC to 0.62.

**Conclusions**

This study established an empirical benchmarking framework for Quantum Kernel Support Vector Machines on imbalanced financial fraud data. When evaluated under normalized sample size and feature constraints, QSVM exhibits classification performance competitive with classical baselines (PR-AUC = 0.730 vs. 0.750). However, unconstrained classical models trained on full transaction datasets remain far superior in runtime, scalability, and predictive accuracy (PR-AUC = 0.983). Future work should explore fault-tolerant quantum hardware execution, variational quantum feature selection, and hybrid quantum neural networks to address current NISQ scaling bottlenecks.

**References:**

1. T. Hastie, R. Tibshirani, and J. Friedman, The Elements of Statistical Learning, 2nd ed. New York, NY, USA: Springer, 2009.
2. L. Breiman, "Random forests," Machine Learning, vol. 45, no. 1, pp. 5–32, 2001.
3. C. Cortes and V. Vapnik, "Support-vector networks," Machine Learning, vol. 20, no. 3, pp. 273–297, 1995.
4. A. Dal Pozzolo, O. Caelen, R. A. Johnson, and G. Bontempi, "Calibrating probability with undersampling for unbalanced classification," in Proc. IEEE Symp. Comput. Intell. Data Mining, 2015.
5. J. Jurgovsky et al., "Sequence classification for credit-card fraud detection," Expert Systems with Applications, vol. 100, pp. 234–245, 2018.
6. V. Havlí d;ek et al., "Supervised learning with quantum-enhanced feature spaces," Nature, vol. 567, pp. 209–212, 2019.
7. M. Schuld and N. Killoran, "Quantum machine learning in feature Hilbert spaces," Physical Review Letters, vol. 122, no. 4, p. 040504, 2019.
8. IBM Quantum, Qiskit Machine Learning Documentation, 2023.
9. E. Farhi and H. Neven, "Classification with quantum neural networks on near-term processors," arXiv preprint arXiv:1802.06002, 2018.
10. M. Cerezo et al., "Variational quantum algorithms," Nature Reviews Physics, vol. 3, pp. 625–644, 2021.
11. A. Dal Pozzolo, O. Caelen, R. A. Johnson, and G. Bontempi, "Credit card fraud detection: A realistic modeling and a novel learning strategy," IEEE Trans. Neural Netw. Learn. Syst., vol. 29, no. 8, pp. 3784–3797, Aug. 2018.

12. J. Preskill, "Quantum computing in the NISQ era and beyond," Quantum, vol. 2, p. 79, 2018.
13. H. Bao and Z. Yang, "A review of SMOTE methods for imbalanced data," IEEE Access, 2023.
14. J. Hong and E. Lopez, "Quantum machine learning: A review of methods and applications," IEEE Access, 2025.
15. M. Grossi, N. Ibrahim, V. Radescu, R. Loredo, K. Voigt, C. Von Altrock, and A. Rudnik, "Mixed quantum-classical method for fraud detection with quantum feature selection," IEEE Transactions on Quantum Engineering, vol. 3, pp. 1–12, 2022.